\documentclass[conference]{IEEEtran}
\IEEEoverridecommandlockouts

\usepackage[T1]{fontenc}
\usepackage[hyphens]{url}

\usepackage[a4paper,top=2cm,bottom=30mm,left=3cm,right=3cm,marginparwidth=1.75cm]{geometry}

\usepackage{amsmath}
\usepackage{graphicx}

\usepackage{cite}
\usepackage{amsmath,amssymb,amsfonts}
\usepackage{algorithmic}
\usepackage{graphicx}
\usepackage{textcomp}
\usepackage{xcolor}
\def\BibTeX{{\rm B\kern-.05em{\sc i\kern-.025em b}\kern-.08em
    T\kern-.1667em\lower.7ex\hbox{E}\kern-.125emX}}
\begin{document}

\title{Sustainable Cloud-Fog-Edge Systems for Smart Buildings: Algorithmic Adaptation under Memory Constraints and Dynamic Orchestration Strategies
}

\author{\IEEEauthorblockN{Mamadou Sow}
\IEEEauthorblockA{\textit{University Sorbonne Paris Nord} \\
\textit{LIPN, UMR CNRS 7030} \\
Villetaneuse, France \\
mamadou.sow@univ-paris13.fr}
\and
\IEEEauthorblockN{Christophe Cérin}
\IEEEauthorblockA{\textit{University Sorbonne Paris Nord} \\
\textit{LIPN, UMR CNRS 7030}\\
\textit{INRIA DATAMOVE}\\
Paris/Grenoble, France \\
christophe.cerin@univ-paris13.fr}}

\maketitle

\begin{abstract}
The escalating environmental footprint of the Information and Communication Technology (ICT) sector—with global Data Center electricity usage potentially reaching 3\% by 2030 ~\cite{iea-AI-Energy-2025} and the imperative for real-time responsiveness in smart buildings demand a foundational shift towards Edge Computing~\cite{AI-generative-X24}. This article proposes a holistic architectural and algorithmic approach for creating autonomous, resilient, and privacy-preserving learning systems within smart buildings, advancing the core concept of the building itself as the data center.
Our contributions, demonstrating originality compared to conventional centralized systems, are structured around three main pillars: 1) Validation of resilient data acquisition chains using the affordable Arduino ecosystem (Uno R4 WiFi/BME280) and secure persistence solutions including centralized archiving (Node-RED/MySQL) and secure remote transfer (Python/rsync/SSH). 2) The proposal of a lightweight, KOptim orchestration framework, interoperable with the FIWARE standard~\cite{lbl}, for the dynamic management of Service Level Agreements (SLA). 3) The adaptation and experimental validation of the Generalized Hebbian Algorithm (GHA) for Principal Component Analysis (PCA) on low-power microcontrollers. We introduce and compare an iterative learning strategy using data buffers, featuring sorting and explicit preservation of extreme samples to maintain diversity. Experimental results on the high-dimensional Tour Perret sensor data confirm that this novel method, despite introducing a necessary temporal complexity of $O(W \log W)$ due to sorting, achieves convergence and final result quality that is visually equivalent to processing the entire dataset in full memory, reliably identifying two distinct operational states. This analysis explicitly incorporates a critical review of the limitations inherent in the Edge-Fog paradigm, detailing security vulnerabilities and the higher power consumption of Fog nodes~\cite{iea2025}.
\end{abstract}

\begin{IEEEkeywords}
Cloud-Fog-Edge\cite{DBLP:conf/vts/Arabi21}, Smart Building, Embedded Artificial Intelligence (Edge AI), GHA-PCA, Microcontroller, FIWARE~\cite{lbl}, Dynamic Orchestration, Algorithmic Complexity, Data Sovereignty
\end{IEEEkeywords}

\section{Introduction}\label{sec:one}


The increasing scrutiny of the ICT sector’s substantial environmental footprint\cite{iot-computing, fog-computing} and the rapid growth of data from IoT devices\cite{service-computing} necessitate a transition from centralized Cloud infrastructures to architectures that support local processing\cite{iot-computing, fog-computing}. This Edge Computing approach, where programs execute close to the request sources for rapid response, is vital for time-sensitive applications like resource optimization in buildings.
Our research promotes the development of inclusive and sustainable technological solutions, particularly with a humanitarian focus, advocating for ecologically responsible, locally deployable systems based on cost-effective microcontrollers. The goal is to design an embedded AI system that operates independently within the building, continuously learning from real-time sensor data\cite{fog_horizons,SANGER1989459}.
This paper details the originality of our work by providing robust comparisons of algorithmic performance and a comprehensive discussion of architectural trade-offs, directly addressing the required critiques.


\section{Architectural Framework and Deployment}\label{sec:two}

\subsection{The Cloud-Fog-Edge Model and Orchestration}

The proposed infrastructure balances computational needs across the Edge (sensors/housing units), the Fog (in-building servers), and the Cloud (central management). This decentralized model supports mobility and offers low latency.

The orchestration framework, KOptim, is a lightweight, Kubernetes-like system designed to manage containers by associating them with dynamic Service Level Agreements (SLAs). The novelty lies in defining SLAs using a range of values instead of fixed quantities. KOptim is coupled with the FIWARE standard to ensure interoperability and to access a policy database for dynamic SLA management. This architecture allows KOptim to negotiate resource allocation with the FIWARE component when Fog nodes are busy

\begin{figure}[hbtp]
    \centering
    \includegraphics[width=0.95\linewidth]{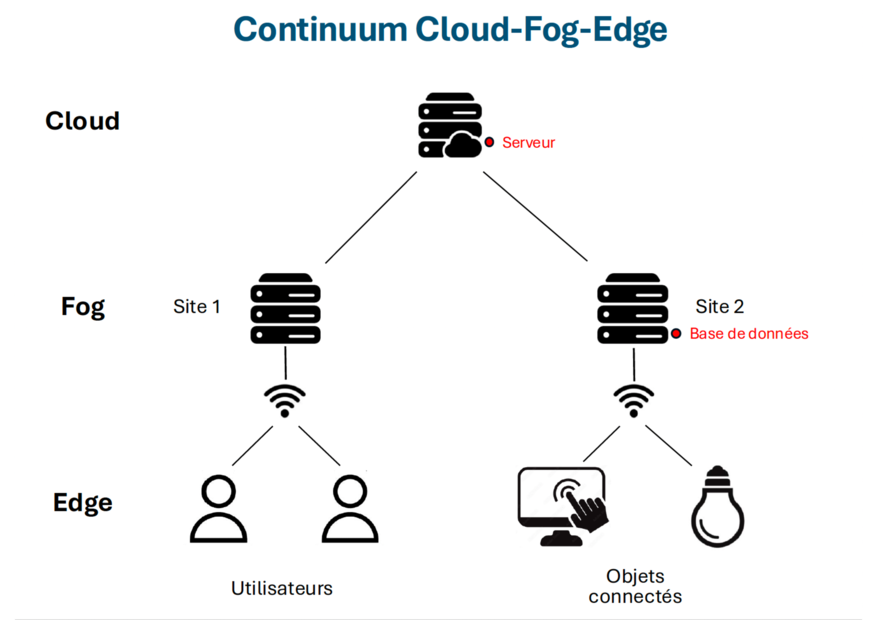}
    \caption{Architectural concerns, illustrating the hierarchy of Cloud, Fog, and Edge levels for smart building management}
    \label{fig:one}
\end{figure}


%
%

\subsection{Inherent Drawbacks of the Decentralized Architecture}

While Edge Computing offers low latency and improved privacy, the architecture presents recognized limitations :
\begin{itemize}
    \item Security Challenges: The complexity of encryption algorithms and security policies can hinder data exchange. Threats such as IP address spoofing, man-in-the-middle attacks, and wireless network vulnerabilities are major concerns. 
    \item Power Consumption: The architecture results in higher power consumption as the Fog nodes are more energy-intensive than a purely centralized Cloud architecture. 
    \item Data Management: Achieving strong data consistency in a distributed, heterogeneous environment is difficult, and managing global storage remains complex despite the design goal. 
\end{itemize}

\begin{figure}[hbtp]
    \centering
    \includegraphics[width=0.95\linewidth]{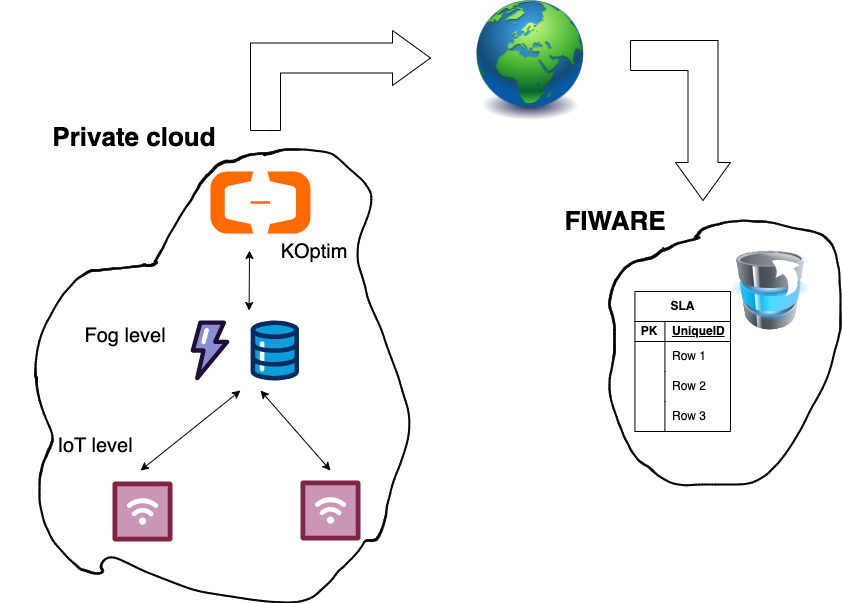}
    \caption{KOptim coupled with FIWARE}
    \label{koptim-arch}
\end{figure}

\section{Methodology I: Data Acquisition and Persistence Chains}\label{sec:trois}

The practical implementations demonstrate modularity and resilience, utilizing the Arduino Uno R4 WiFi and cost-effective sensors.

\subsection{Embedded Web Server for Real-Time Monitoring}


The Arduino Uno R4 WiFi hosts a web server.

\begin{itemize}
    \item Acquisition: The BME280 sensor reads Temperature, Humidity, Pressure, and estimated Altitude via I2C. 
    \item Web Page Generation: The firmware dynamically generates an HTML page stored in PROGMEM. Placeholders (e.g., \%TEMP\_PLACE\_HOLDER\%) are replaced with real-time sensor values. The page includes a \texttt{<meta http-equiv="refresh" content="5">} tag for automatic refresh. Concurrently, the system outputs a continuous JSON stream to the serial monitor for logging. 
\end{itemize}

\begin{figure}[hbtp]
    \centering
    \includegraphics[width=0.9\linewidth]{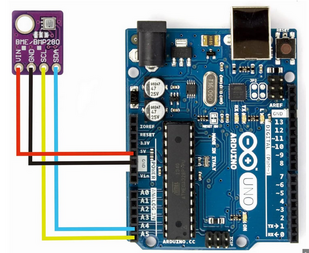}
    \caption{Wiring diagram for the BME230 i2c 3.3V sensor on the Arduino Uno R4 Wifi MCU}
    \label{fig:two}
\end{figure}

\begin{figure}[hbtp]
    \centering
    \includegraphics[width=1\linewidth]{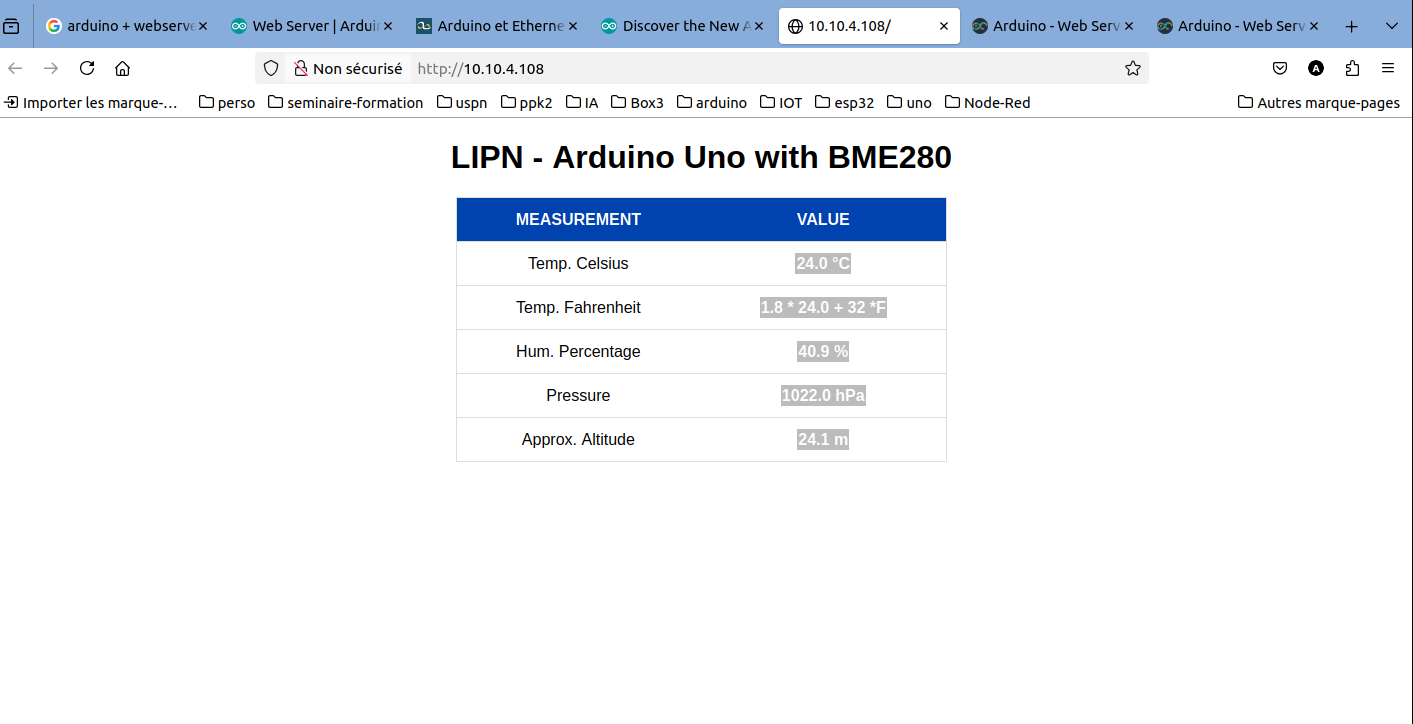}
    \caption{Content displayed in the Web Browser: The displayed URL is the IP address 10.10.4.108}
    \label{fig:three}
\end{figure}

\subsection{Secure Centralized Persistence via Node-RED}

For long-term archival, the Node-RED platform acts as the central server.
\begin{itemize}
    \item Data Flow: The Arduino transmits the four readings (T, H, P, A) as a CSV string over the USB-C serial port. 
    \item Analysis: A Node-RED Function Node parses the CSV string, converts elements to floating-point numbers, and structures the data into a JSON object. 
    \item Security and Insertion: To guarantee robustness and prevent SQL injection attacks, the insertion SQL query is dynamically constructed using placeholders (?) in the msg.topic property. The msg.payload property is defined as an ordered array corresponding to the values. This approach ensures the integrity of the records in the MySQL database (meteo\_iot).
\end{itemize}

\subsection{Secure Remote Transfer and Consolidation}

For resilient remote storage, a Python script automates data logging and transfer.
\begin{itemize}
    \item Local Logging: The Arduino sends CSV data (including an internal timestamp in milliseconds) at a high baud rate (115200). The Python script adds the real PC timestamp for traceability and saves it locally (data.csv). 
    \item Periodic Secure Transfer: The script periodically transfers the local file to a remote server using rsync via SSH, leveraging existing computer security protocols. 
    \item Remote Consolidation: After a TOTAL\_DURATION (e.g., 600 seconds), the script connects remotely to aggregate all individual, timestamped files (data\_*.csv) into a single, consolidated file (consolidated\_data.csv), managing headers to prevent duplication.
\end{itemize}

\begin{figure}[hbtp]
    \centering
    \includegraphics[width=0.8\linewidth]{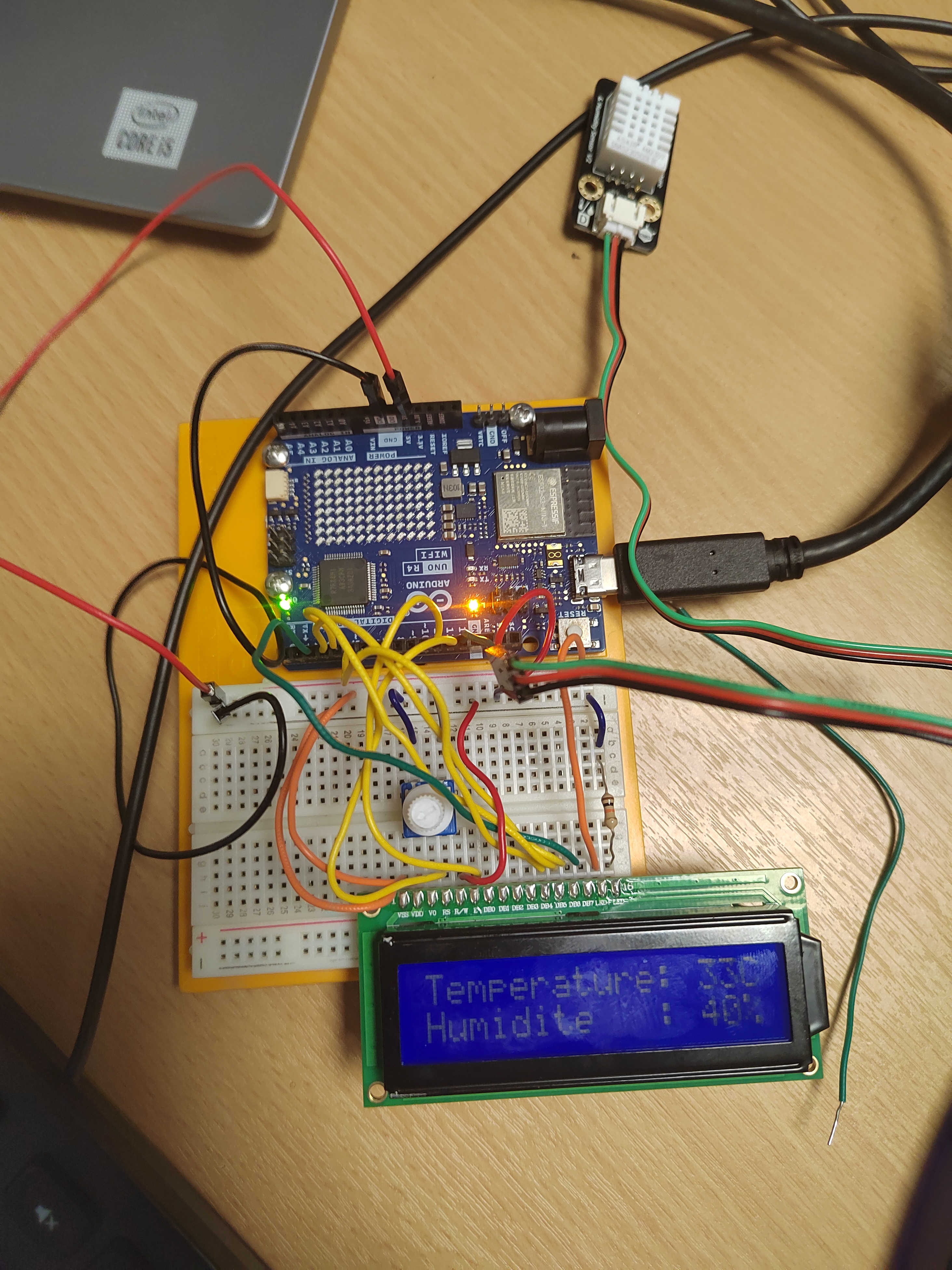}
    \caption{Electronic Circuit}
    \label{fig:four}
\end{figure}



\section{Methodology II: Embedded Artificial Intelligence (Edge AI)}

The design of memory-efficient algorithms for low-power microcontrollers (Arduino ecosystem) is critical.


\subsection{GHA-PCA for Dimensionality Reduction}


We implement the Generalized Hebbian Algorithm (GHA) in C++ using the Eigen library to perform Principal Component Analysis (PCA) iteratively. This method is chosen because it avoids calculating the memory-intensive covariance matrix. The algorithm is applied to the multidimensional sensor data from the Tour Perret building, specifically using normalized features (columns 1, 3, 5, 6: accMotion, humidity, temperature, and vdd). The weights $W$ converge toward the eigenvectors of the data covariance matrix.

\subsection{Comparative Strategies for Limited Memory (Original Contribution)}


To simulate the constrained environment of an embedded device (e.g., limited RAM) and allow the algorithm to process large datasets sequentially, we compared three learning strategies:
\begin{enumerate}
    \item Reference Method (Full Memory): Processing the entire dataset at once (infeasible on microcontrollers). 
    \item Simple Block Processing: Iterative processing using fixed-size data buffers $W$ (e.g., 1024 lines). This simulates diversity by simply moving to a new block. 
    \item Block Processing with Explicit Diversity Management (Proposed Method): This innovative strategy explicitly manages the buffer content to ensure diversity.
    \begin{itemize}
        \item The buffer is filled. 
        \item The buffer is sorted (e.g., based on accMotion value), complexity $O(W \log W)$. 
        \item Only samples at the extremities of the distribution (e.g., the lower and upper halves) are retained to maximize sample diversity. 
        \item The resulting empty spaces are refilled with new data. 
        \item The GHA step is executed on this diversified buffer.
    \end{itemize}

\end{enumerate}

\subsection{Time Complexity Analysis}

The inclusion of the sorting step in the proposed method increases the temporal complexity of each iteration. The complexity is the sum of the GHA calculation and the sorting operation, resulting in an added cost of $O(W \log W)$ per buffer cycle. This complexity is justified by the necessity of preserving model quality and ensuring stable convergence in memory-constrained settings.

Looking at the table in Figure \ref{fig:three}, we can say that the ESP32 microcontroller can easily receive a stream of 1024 messages with optimized batch processing of 128 messages.

\section{Results and Comparative Validation}



\begin{figure}[hbtp]
    \centering
    \includegraphics[width=0.95\linewidth]{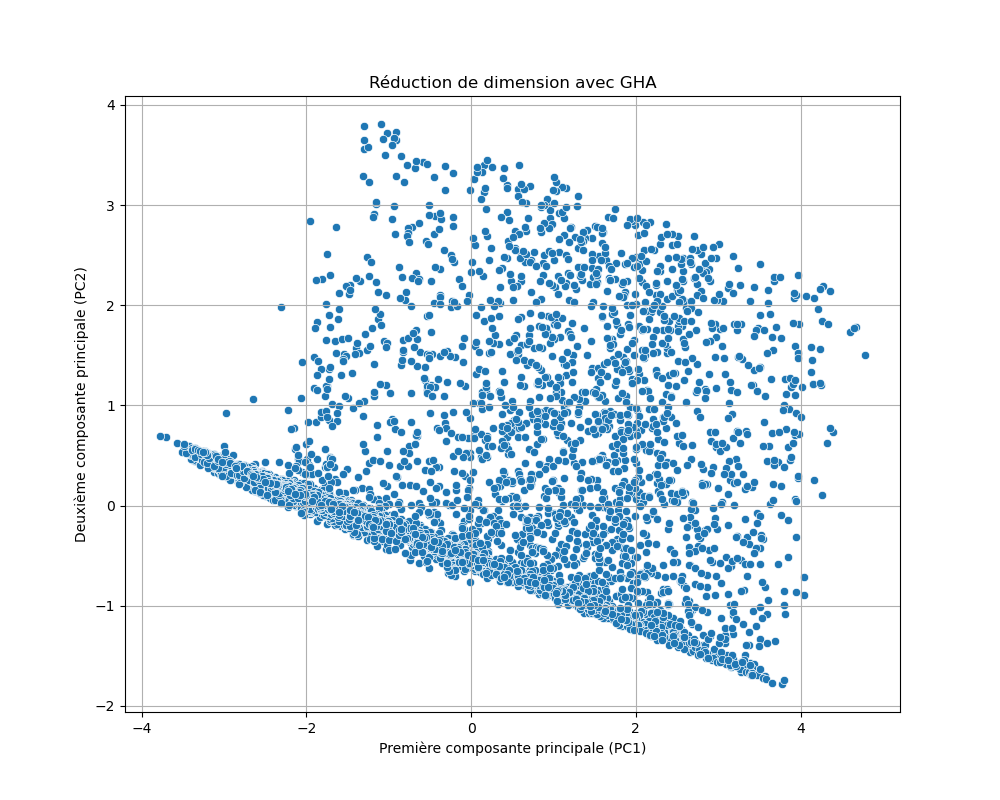}
    \caption{Scatter Plot illustrating Dimension reduction with GHA (Block Processing with Diversity Management)}
    \label{fig:eleven}
\end{figure}

\subsection{GHA-PCA Convergence Quality and Originality}

We analyze the projections onto the principal components (PC1 and PC2). To effectively visualize data density and avoid overplotting, we primarily use Kernel Density Estimation (KDE) plots.
\begin{itemize}
    \item Simple Block Processing: The resulting clusters appear slightly less concentrated and more dispersed compared to the reference, indicating a less stable convergence in capturing the fine structure of the data. 
    \item Block Processing with Diversity Management: This result is visually the closest to the reference figure. The density of the two clusters is highly concentrated with more distinct edges. This demonstrates the effectiveness and originality of the sorting strategy in maintaining sample diversity and enabling the algorithm to generalize and converge towards the optimal solution. 
\end{itemize}

\begin{figure}[hbtp]
    \centering
    \includegraphics[width=1\linewidth]{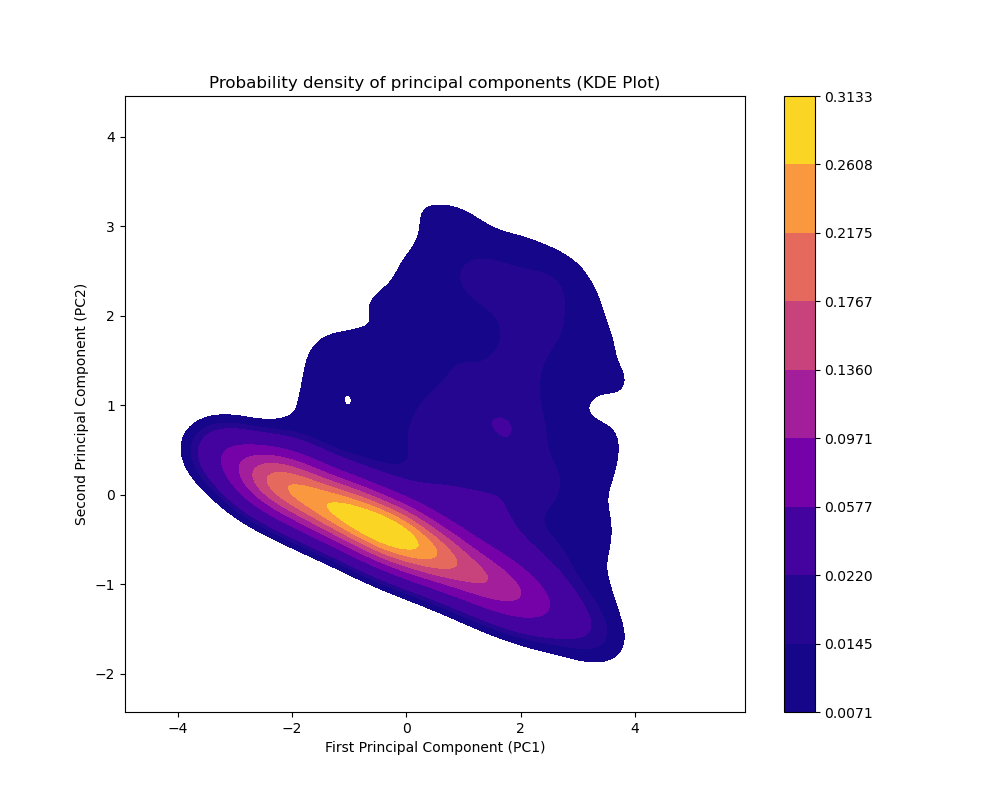}
    \caption{Probability density of principal components (KDE Plot)}
    \label{fig:seven}
\end{figure}

\textbf{Structural Interpretation}

The analysis robustly identifies the existence of two distinct and dominant operating states or clusters in the sensor data. The First Principal Component (PC1) captures the greatest variance and acts as the primary axis separating these two states.

\subsection{K-MEANS Clustering for Activity Analysis}

\subsubsection{Data mining from the Perret Tower with the GHA dimension reduction algorithm}


This study focuses on applying a dimensionality reduction algorithm to a dataset originating from the Tour Perret building in Grenoble. Dimensionality reduction involves the transformation of data from a high-dimensional space into a lower-dimensional space, ensuring that the reduced representation retains essential characteristics of the original data, ideally approximating its intrinsic dimension. The Tour Perret dataset comprises up to 15 dimensions. Our objective is to achieve a two-dimensional representation to facilitate more effective visual analysis. In this context, we demonstrate the implementation of the GHA-PCA algorithm\cite{SANGER1989459}.

We now discuss the adaptation of the abstract algorithm to fit our requirement to run on embedded devices and, more importantly, to transform it into an online algorithm.

The developed code utilizes the Eigen library for matrix operations, which is a popular choice in C++ for linear algebra tasks. This C++ code retains the core functionality of the original Python code while adapting it to the syntax and conventions of C++. The program will process the data in blocks of 1024 lines of the Perret Tower data ~\cite{TourPerrethead10col.csv}, display the iteration number at each step, and update the eigenvalues and eigenvectors accordingly. In the end, the results will be saved in a CSV file and displayed graphically.
 
\paragraph{Explanation} We have introduced a constant W that defines the size of the buffer (1024 lines). The main loop now runs through the data in blocks of 1024 lines. At each iteration, we display the current iteration number and the lines currently being processed. For each block of 1024 lines, we apply the GHA algorithm on each line of the block. If the total number of lines is not a multiple of 1024, the last block will be smaller. We use $\min(W, n-i)$ to handle this. The Eigen template library already manages memory dynamically.

The header file ghapca.h contains the declarations of variables, constants, and shared functions needed for the source code\footnote{\url{https://github.com/madou-sow/OnlineML_ESP32/blob/main/ARDUINO/GHA-PCA/src/online_GhaPca_update_buffer.cpp}} to be compiled into file. In a second program\footnote{\url{https://github.com/madou-sow/OnlineML_ESP32/blob/main/ARDUINO/GHA-PCA/src/online_GhaPca_update_buffer_normalize.cpp}}, there is a special feature, namely, a section where the data is normalized. This practice allows for transforming the data without distorting it. Normalization consists of harmonizing the data so that all the entries of the different data sets that relate to the same terms follow a similar format.
    
\paragraph{Analysis} The differences between the results in C++ and Python\footnote{\url{https://github.com/madou-sow/OnlineML_ESP32/blob/main/MICROPYTHON-ULAB/DimensionalityReduction/testing_batch_ghapca_json.py}} programs can be attributed to several factors. The C++ program processes the data in blocks of 1024 rows, while the Python program processes the data sequentially. This may affect how the eigenvalues and eigenvectors are updated. The initial parameters, such as the eigenvalues and the eigenvectors, may be initialized differently in the two programs, which may influence the final results. The C++ program uses an online approach to update the eigenvalues and eigenvectors, while the Python program uses a batch approach. This may lead to differences in the convergence of the results. The Python program ignores rows without payload, which can reduce the number of data rows processed, while the C++ program does not seem to have this logic. The C++ program normalizes the data by ignoring some columns (1, 3, and 6), while the Python program normalizes all columns. This can lead to differences in the values of the processed data. The possible causes are i) Constant column: Columns 1, 3, and 6 may contain a constant value (e.g., all elements are 0, 1, or some other fixed value).; ii) Missing or poorly formatted data: If columns 1,3, and 6 contain missing or poorly formatted data, it could lead to a situation where all values are the same after processing; iii) Data loading issue: The CSV file might not be loaded properly, resulting in incorrect values in columns 1, 3, and 6.
    
\paragraph{How to bypass the previous difficulties related to data quality} We implemented the following solutions. If columns 1, 3, and 6 are not important for your analysis, you can ignore them when normalizing. If columns 1, 3, and 6 are important but contain constant values, you can replace these values with a default value or a small variation to allow normalization. If normalization is a problem, you can consider using standardization (subtract the mean and divide by the standard deviation) instead of normalization. This works even if the values are constant (although the standard deviation is zero in this case, which would also require special handling).

\subsubsection{Online machine learning and Online technique with CampusIOT's LoRaWAN Datasets and the K-means Clustering Algorithm}

Again, in this subsection, our goal is to design a model that can learn from new data without revisiting past data, a model that is also robust to concept drift. The concept of drift refers to the evolutionary changes in the data that render an existing data model obsolete. This phenomenon occurs when there are unforeseen alterations in the statistical properties of the target variable that the model is designed to predict over time. The data set used also comes from the Smart Building ecosystem. This developed model will be very close to what happens in a production context, which is generally event-based. It will integrate well with the rest of the embedded systems programming ecosystem.

In the simulation, the CO2 concentration and temperature are published from our local MQTT server with the subjects "C02/Celsius" and "final/final". The data is received via the ESP32 board and is used by the Arduino IDE, where a program~\cite{esp32MqttKmeansClusterQuickBuffer} has been implemented to perform clustering with the K-MEANS algorithm. In our case, we considered a structure as a means of defining this type of data, although it is more than a simple container for primitive data types because we can also define other functionalities. What it does with the elements or data points is a movement that meets the criteria at the end of the range. It then returns an iterator to the first element of the deleted elements (which are just moved).

The objective is to analyze the behavior of the ESP32 controller in the face of a stream of messages received continuously, subject to an implemented algorithm containing both a clustering algorithm, which makes it possible to identify a maximum of 4 clusters.

During the experiments, we managed to send several message streams (256, 512, 1024, and 5120) containing CO2 quantities and temperatures from the MQTT server to the ESP32 microcontroller. Using our integrated Arduino development environment, we have carried out a distribution of the data to be processed by the microcontroller. This variable is called W. It varies from 8 to 512. 512 is the maximum value taking into account the saturation that can occur.

By referring to an equation with 3 unknowns, namely x (set of messages), y (splitting or subset of messages), and z (number of splits of messages processed), the satisfactory result corresponds to the case allowing the observation of a maximum of messages processed in a message set.

\begin{figure}[hbtp]
    \centering
    \includegraphics[width=0.95\linewidth]{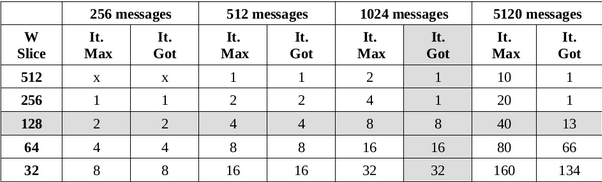}
    \caption{Message Flow, Processing Slice and Number of Possible Iterations}
    \label{fig:twelve}
\end{figure}

Looking at the table in Figure \ref{fig:three}, we can say that the ESP32 microcontroller can easily receive a stream of 1024 messages with optimized batch processing of 128 messages.

\subsubsection{KMeans Clustering for the CampusIOT's LoRaWAN Datasets}

We consider here the "ELSYS-ERS-CO2, ERS-CO2-48B30 W=512, Data = 5120, K=4" use case, which is the settings. This input file contains the basic information (ID, CO2, T(°C), Cluster) about the data recorded. Clustering involves dividing points in a data set into several groups such that data points in the same group are more similar to other data points in the same group than to those in other groups. The goal in clustering is to separate groups with similar characteristics and distribute them into clusters.

We define a target number k, which refers to the number of centroids we need in the dataset. As for the centroid, it is the imaginary or real location representing the center of the cluster. Each point in the dataset is assigned to each of the clusters by reducing the sum of squares within the cluster. Alternatively, the K-means algorithm identifies k number of centroids and then allocates each data point to the closest cluster, while keeping the centroids as small as possible. The “means” in K-means refers to find the center of gravity of the data. Using the Seaborn module as well as the Matplotlib module, we will represent the distribution diagram with different variations of the data, such as cluster1data5120w512.csv.  A distribution diagram in the Figure \ref{fig:four} represents the variation in the distribution of data by a histogram and a line in combination with it. 

\begin{figure}[htbp]
    \centering
    \includegraphics[width=0.95\linewidth]{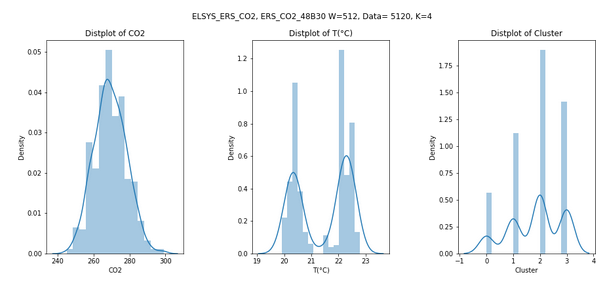}
    \caption{Distribution Diagram ELSYS\_ERS\_CO2, ERS\_CO2\_48B30 W=512, Data= 5120, K=4}
    \label{fig:nine}
\end{figure}

\section{Critical Analysis and Limitations}\label{sec:four}


\subsection{Inherent Limitations of the Proposed Methodology}

We detail the disadvantages and limitations of the proposed approach, as required by the critiques (Critique 2):
\begin{itemize}
    \item PCA Linearity Limit: PCA (and GHA) is a linear method. It can only capture linear relationships between variables. If data relationships are non-linear, this method may not capture the true structure. 
    \item Information Loss: Dimensionality reduction entails an inevitable loss of total variance. 
    \item Implementation and Performance Cost: The explicit memory management strategy (sorting) increases the implementation complexity and the time complexity to $O(W\log W)$ per iteration, which affects speed on low-power devices. 
    \item Dependency on Sorting Criterion: The success of the diversity preservation method relies heavily on choosing the correct variable for sorting (e.g., accMotion). A poor choice could compromise the preservation of diversity. 
    \item Ecosystem Compatibility: There is a recognized time lag in algorithm development between microcontroller ecosystems (ARDUINO and Micropython). 
\end{itemize}

\section{Conclusion and future works}\label{sec:five}


We have presented an integrated and critically validated approach for sustainable Cloud-Fog-Edge systems in smart buildings. This work confirms the technical feasibility of Edge Intelligence on cost-effective microcontrollers, demonstrating that optimized iterative learning strategies, such as GHA-PCA with explicit diversity management, can overcome severe memory constraints while maintaining high analytical quality. The robust identification of two distinct operating modes in the sensor data validates the effectiveness of this adaptation for analyzing hidden building dynamics.

The architectural backbone—ranging from autonomous embedded acquisition (Arduino Uno R4 WiFi/BME280) and secure persistence (Node-RED/MySQL) to dynamic orchestration (KOptim/FIWARE)—establishes a resilient, decentralized ecosystem.
We propose treating the building itself as the data hub, adhering to edge computing concepts to protect privacy and lessen dependence on external cloud services. 

This is especially important in humanitarian settings, where data ownership, energy conservation, and infrastructure limitations are crucial. We describe a lightweight orchestration framework akin to "Kubernetes" for the deployment of AI services within these environments and exhibit our advances in applying AI algorithms on low-power, affordable microcontrollers, such as those in the Arduino ecosystem. By facilitating on-site learning directly on sensors or microcontrollers, our initiative aims to introduce intelligent services to resource-constrained settings, enhancing autonomy, resilience, and sustainable growth in vulnerable or underserved communities.

Future work must address the challenges related to the delay in algorithm advancements between microcontroller ecosystems, the necessity of further validation across \textbf{additional datasets}, and a comprehensive comparative assessment of the global energy efficiency of this decentralized architecture relative to traditional Cloud solutions.


\bibliographystyle{IEEEtran}
\bibliography{IEEEabrv,mybib}

\end{document}